\pdfoutput=1

\documentclass[11pt]{article}

\usepackage[preprint]{acl}

\usepackage{times}
\usepackage{latexsym}
\usepackage{booktabs}
\usepackage{multirow}
\usepackage{xtab}
\usepackage{array}
\usepackage{kotex}

\usepackage[T1]{fontenc}

\usepackage[utf8]{inputenc}

\usepackage{microtype}

\usepackage{inconsolata}

\usepackage{graphicx}
\usepackage{xspace}
\usepackage{xcolor}
\usepackage{enumitem}
\usepackage{amsmath,amssymb}

\usepackage{colortbl}
\usepackage{tcolorbox}

\usepackage{cleveref}

\crefname{figure}{Fig.}{Figs.}
\crefname{equation}{Eqn.}{Eqns.}
\crefname{appendix}{Appx.}{Appx.}
\crefname{table}{Table}{Tables}
\crefname{section}{\S}{\S\S}

\newcommand{\gpt}{GPT-4o\xspace}

\newcommand{\llama}{Llama-3\xspace}
\newcommand{\qwen}{Qwen-2.5\xspace}

\newcommand{\llamasmall}{\texttt{Llama-3.1-{8B}}\xspace}
\newcommand{\llamalarge}{\texttt{Llama-3.3-{70B}}\xspace}
\newcommand{\gptsmall}{\texttt{gpt-{4o}-mini}\xspace}
\newcommand{\gptlarge}{\texttt{gpt-{4o}}\xspace}
\newcommand{\qwensmall}{\texttt{Qwen-2.5-{14B}}\xspace}
\newcommand{\qwenlarge}{\texttt{Qwen-2.5-{72B}}\xspace}
\newcommand{\gemmasmall}{\texttt{gemma-2-{9b}}\xspace}
\newcommand{\gemmalarge}{\texttt{gemma-2-{27b}}\xspace}
\newcommand{\ayasmall}{\texttt{aya-exp-{8b}}\xspace}
\newcommand{\ayalarge}{\texttt{aya-exp-{32b}}\xspace}
\newcommand{\geminilarge}{\texttt{gemini-2.0}\xspace}
\newcommand{\geminismall}{\texttt{gemini-2.0-lite}\xspace}

\newcommand{\discrepancy}{discrepancy}

\definecolor{asparagus}{rgb}{0.53, 0.66, 0.42}

\title{Found in Translation: Measuring Multilingual LLM Consistency\\as Simple as Translate then Evaluate}

\author{
\begin{tabular}{c}
Ashim Gupta \quad Maitrey Mehta \quad Zhichao Xu \quad Vivek Srikumar
\end{tabular}\\
Kahlert School of Computing \\
University of Utah \\
\texttt{ashim@cs.utah.edu}}

\begin{document}
\maketitle

\begin{abstract}
Large language models~(LLMs) provide detailed and impressive responses to queries in English. However, \textit{are they really consistent at responding to the same query in other languages?} The popular way of evaluating for multilingual performance of LLMs requires expensive-to-collect annotated datasets. Further, evaluating for tasks like open-ended generation, where multiple correct answers may exist, is nontrivial. 
Instead, we propose to evaluate the predictability of model response across different languages. 
In this work, we propose a framework to evaluate LLM's cross-lingual consistency based on a simple \textit{Translate then Evaluate} strategy. 
We instantiate this evaluation framework along two dimensions of consistency: information and empathy. 
Our results reveal pronounced inconsistencies in popular LLM responses across thirty languages, with severe performance deficits in certain language families and scripts, underscoring critical weaknesses in their multilingual capabilities.
These findings necessitate cross-lingual evaluations that are consistent along multiple dimensions. We invite practitioners to use our framework for future multilingual LLM benchmarking.    
\end{abstract}
\section{Introduction}
\label{sec:introduction}
Large language models (LLMs) have demonstrated impressive capabilities on English-language tasks. 
Most popular LLMs like \gpt, \llama, \qwen, etc. achieve performances north of 85\% on popular benchmarks like MMLU, HellaSwag, WinoGrande, etc. for English~\citep{yang2024qwen2}. 
Increasingly, these models are now deployed in multilingual, multicultural contexts, and are marketed as capable multilingual systems~\citep{meta2024ai}.
However, understanding and characterizing their performance profiles in different languages remains a critical and open challenge.
%
%
This difficulty is partly due to the high cost of scaling annotation efforts across many languages. Moreover, even with ample resources to support annotation, evaluation remains difficult for open-ended tasks like long-form question answering or dialogue generation, where a single correct answer may not exist. 
\begin{figure*}[ht]
    \centering
    \includegraphics[width=\linewidth]{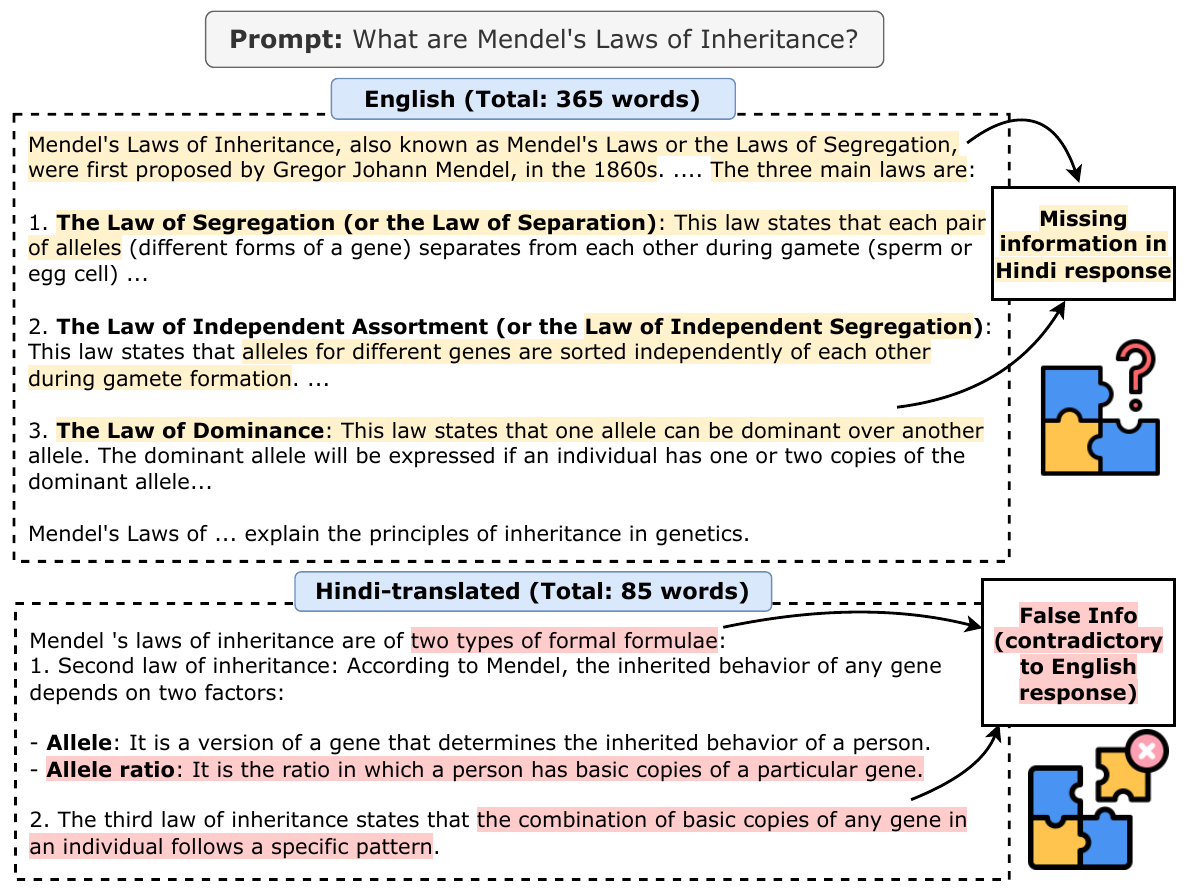}
    \caption{Llama-3 8B~(\raisebox{-0.2em}{\includegraphics[height=1em]{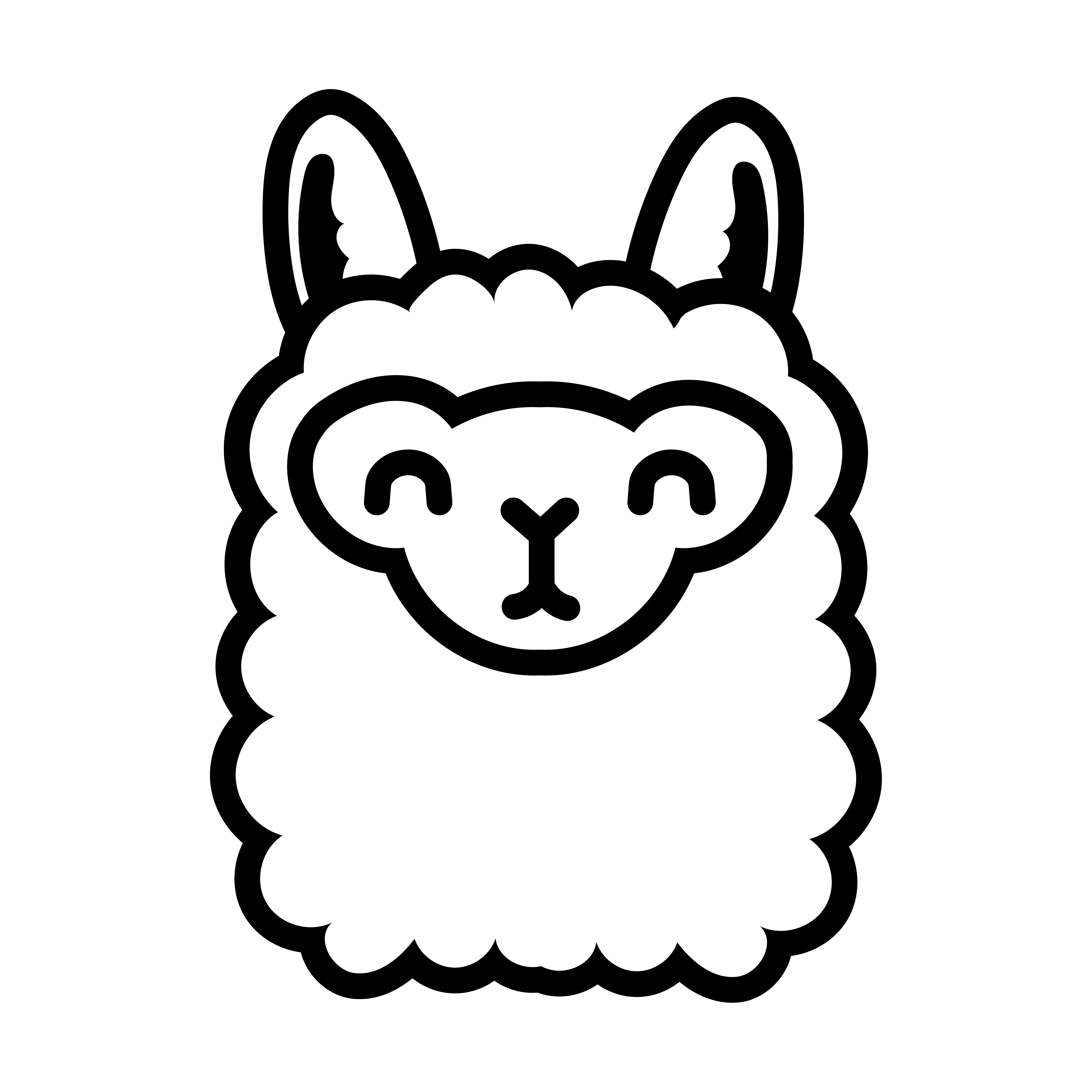}}) response to an English~(\texttt{en}) prompt versus its response to the translation of the same prompt in a different language like Hindi~(\texttt{hi}) .  We observe a clear gap in information generated between the two languages. The responses in Hindi has been translated for the reader. The English response is truncated for brevity.}
    \label{fig:intro_motivating_example}
\end{figure*}

In this paper, we propose a framework to study multilingual performance profiles of LLMs by assessing their \textit{cross-lingual consistency}\,---\,comparing model outputs in various languages against their corresponding English responses. Our framework \emph{sidesteps the need for expensive annotated data and the formulation of close-form tasks}.
Cross-lingual consistency matters for two key reasons. First, consistency is inherently valuable in model behavior, as noted in prior work~\cite{gardner-etal-2020-evaluating, ribeiro-etal-2020-beyond}. A system that behaves consistently across languages is more predictable, which is especially critical in high-stakes, user-facing domains such as healthcare or legal services~\citep{schlicht2025llms}. Second, consistency ensures equitable access to information.
Consider the example in~\cref{fig:intro_motivating_example}; a scientific question posed in Hindi elicits a response that is both shorter and less informative than its English counterpart.


It is challenging to evaluate consistency on open-ended tasks like open-domain question answering. For close-form tasks like classification or multi-choice question answering, consistency can be quantified through agreement rates or accuracy. 
For open-ended tasks, ground-truth data is usually absent, and when present, does not encompass all the correct possibilities. In such cases, model-based evaluation metrics are often adopted\,---\,such as FActScore~\cite{min2023factscore}\,---\,which require accurate identification of claims and verification of factual entailment. 
Crucially, such evaluators are optimized for English and a few high-resource languages, making them less reliable when applied to responses in low-resource languages such as Gujarati or Vietnamese. 

To tackle this challenge, we adopt a simple yet effective strategy: \textit{Translate then Evaluate}. We translate LLM responses to non-English queries into English using a high-fidelity translation system, allowing for direct comparison between this translated response and the response to its corresponding English query. 
We use state-of-the-art translator to enable a scalable evaluation of cross-lingual consistency, making it feasible to include a much broader spectrum of languages in our subsequent analyses.

We instantiate our framework across two disparate consistency dimensions: \textit{information consistency}, to study the information overlap in the responses for the same information-seeking prompt across multiple languages, and \textit{empathy consistency}, to study the persistence of mechanisms of empathy for the same therapeutic prompts across different languages. We adopt off-the-shelf English-based classifiers to act as our consistency evaluators for both these dimensions, with intermediate translation systems facilitating language compatibility. 

Our findings show low rates of consistency along both dimensions when tested on mainstream LLMs for 30 languages. We observe that models are generally less consistent for certain language families, scripts and, generally, for languages not seen in the training data. Further, stronger proprietary models are more consistent compared to open-weight models. These findings underscore weaknesses in mainstream LLMs' multilingual capabilities, necessitating a more rigorous, comprehensive multi-dimensional consistency evaluation.
We release our code and data at \url{https:www.github.com/anonymized}. We will release the code as a \texttt{pip} installable package.
\section{A Framework for Cross-lingual Consistency} 
\label{sec:framework}
Evaluating cross-lingual consistency of LLMs is uniquely challenging, especially for open-ended or nuanced tasks where standard accuracy metrics are inapplicable. A key obstacle is that most modern evaluators\,---\,be it for factuality, empathy, or even coherence\,---\,are monolingual and validated primarily on English data. Multilingual evaluation is either unreliable or infeasible in the absence of high-quality resources for other languages.

We address this challenge via a general and scalable framework for measuring cross-lingual consistency, centered around a simple yet effective principle: \textit{Translate then Evaluate}.\footnote{This approach is similar to the widely used and highly effective baseline of translating the input before answering, as noted in~\citet{shi2023languagemodelsaremultilingualcotreasoners}.} This paradigm allows us to exploit well-established English-language evaluation pipelines, extending their utility to a much broader set of languages.

Let us formalize this idea. 
We begin by defining the notion of an evaluator. An evaluator $\mathcal{E}$ is a function that takes two text inputs and returns a compatibility score.  Evaluators may be based on traditional heuristic metrics, such as BLEU~\cite{Papinenibleuevaluationmetric} or ROUGE~\cite{lin-2004-rouge}, which compute n-gram overlap, or on more recent LLM-based evaluation metrics like FActScore~\citep{min2023factscore} and VeriScore~\citep{song2024veriscore}. 

%
Let $\mathcal{X}$ denote a set of abstract prompts, where each prompt $x \in \mathcal{X}$ represents a language-agnostic intent or input (e.g., a question or instruction). For language $\ell \in \mathcal{L}$, we denote by $x_\ell$ the realization of $x$ in language $\ell$. In particular, $x_{\texttt{en}}$ refers to the English version of $x$. We assume that all prompts in $\mathcal{X}$ have valid and semantically equivalent translations in each language $\ell \in \mathcal{L}$.
Additionally, for a given multilingual LLM $\mathcal{M}$, we define its generated response to prompt $x$ in language $\ell$ as:
\begin{align}
\notag
r_{\ell}(x) = \mathcal{M}(x_{\ell}).
\end{align}
%
We denote consistency of the model $\mathcal{M}$ for language $\ell$ with respect to an evaluator $\mathcal{E}$, as $\mathcal{C}_{\mathcal{M},\mathcal{E}}(\ell)$. This consistency is defined as the expected compatibility score computed by an evaluator $\mathcal{E}$ between the English response and non-English response, $r_{\text{en}}(x)$ and $r_{\ell}(x)$. The expectation is computed over prompts $x$, which gives us:
\begin{align}
\notag
\mathcal{C}_{\mathcal{M}, \mathcal{E}}(\ell) := \mathbb{E}_{x}\left[\mathcal{E}\left(r_{\text{en}}(x), r_{\ell}(x) \right)\right]. \label{eq:consistency-definition}
\end{align}
Directly comparing $r_{\text{en}}(x)$ and $r_{\ell}(x)$ is non-trivial; most model-based evaluators are monolingual and optimized for English. Therefore, we use a translation model $\mathcal{T}_{\ell \rightarrow {\text{en}}}$ that maps responses from the target language $\ell$ to English:
\begin{align}
\notag
\hat{r}_{\ell}(x) :=  \hat{r}_{\ell \rightarrow \text{en}}(x) = \mathcal{T}_{\ell \rightarrow {\text{en}}}(r_{\ell}(x))
\end{align}
%
This translation step brings both responses into the same language, allowing the use of English-only evaluators. 
%
%
Assuming $\mathcal{E}$ produces similar scores before and after translation, we can write the consistency as:
\begin{align}
\notag
\mathcal{C}_{\mathcal{M}, \mathcal{E}}(\ell) \approx \mathbb{E}_x\left[\mathcal{E}\left( r_{\text{en}}(x), \hat{r}_{\ell}(x) \right)\right]
\end{align}
%
%
%
Practically we cannot compute the expectation over all possible inputs $x$, therefore we approximate the consistency score as the empirical mean over a set of prompts $x \in \mathcal{X}$:
\begin{equation}
\mathcal{C}_{\mathcal{M}, \mathcal{E}}(\ell) \approx \frac{1}{|\mathcal{X}|} \sum_{x \in \mathcal{X}} \mathcal{E}\left( r_{\text{en}}(x), \hat{r}_{\ell}(x) \right)
\label{eq:consistency}
\end{equation}
%
%
%
The above framework is fully defined by defining the translation model and an (English) evaluator. We now offer some discussion about how each of these might be chosen. 

\paragraph{Desideratum for the translation model.} 
As noted above, a critical requirement of this framework is that the translation model $\mathcal{T}$ preserves the semantic and pragmatic properties that the evaluator $\mathcal{E}$ is designed to measure. In our experiments, we employ the state-of-the-art NLLB 54B-parameter Mixture-of-Experts (MoE) model~\cite{costa2022no}, which has demonstrated superior performance over conventional systems such as Google Translate for English-targeted translation across a wide range of languages.

\paragraph{Evaluators.} 
To assess the consistency of multilingual LLMs, it is important to use a variety of evaluators. Our proposed framework is evaluator-agnostic and can be instantiated with any that support English.
Ideally, a broad suite of evaluators\,---\,each targeting a specific linguistic or functional property\,---\,will help characterize the multilingual performance profiles of LLMs. The evaluator itself may operate in one of the two modes. In the first, it can directly compare a pair of texts to produce a similarity or compatibility score similar to FActScore. In the second mode, it can target a specific property (e.g., sentiment), compute this property independently for each text, and then compare the resulting attributes to derive a score. 
In this work, we instantiate our framework using two readily available, off-the-shelf evaluators operating in these two modes, to be detailed in the next section.
\section{Instantiations of Cross-Lingual Consistency}
As noted earlier (\cref{sec:framework}), the proposed framework does not depend on the specific choice of an evaluator. We discuss two instantiations of this framework for scenarios where high-quality model-based evaluators are not available for languages other than English. Other instantiations of the framework could be similarly applied (e.g., toxicity, sentiment, etc.).

\subsection{Information Consistency}
\label{subsec:info_consistency}
%
%
%
%
\paragraph{Motivation.} For applications involving scientific or factual content, it is critical that multilingual language models produce consistent information regardless of the language in which a question is posed. Inconsistent outputs can lead to unequal access to reliable knowledge across languages, undermining trust in the model. As illustrated in~\cref{fig:intro_motivating_example}, the same scientific question asked in English and Hindi yields responses that differ not only in length but also in factual accuracy\,---\,the Hindi response is both incomplete and scientifically incorrect. In this subsection, we formalize the notion of cross-lingual information consistency to capture and quantify such discrepancies. 
%
\paragraph{Notation and setup.}Given that most state-of-the-art LLMs exhibit stronger performance in English compared to other languages~\citep{grattafiori2024llama}, it is natural to use the English response as a reference when measuring information consistency. To this end, we leverage FActScore~\citep{min2023factscore}, an off-the-shelf model-based evaluator designed to assess the factual accuracy of generated text against a reference. FActScore fits seamlessly into our framework, as it enables us to evaluate the factual consistency of non-English responses with respect to the information conveyed in the English output. Specifically, FActScore represents the factual precision of a generated response $r$ with respect to a knowledge source $s$:
\begin{equation}
\mathcal{E}_{\text{FAct}}(r, s) = \frac{1}{|F(r)|} \sum_{f \in F(r)} \mathbf{1}_{\text{supported}(f, s)}
\end{equation}
where $F(r)$ denotes the set of  claims or pieces of information extracted from the response $r$, and $\text{supported}(f, s)$ indicates whether claim $f$ is supported by the reference text $s$. In our setup, we use the English response $r_{\texttt{en}}(x)$ as the reference, and evaluate the translated non-English response $\hat{r}_{\ell}(x)$ for factual consistency (substituting into~\cref{eq:consistency}):
\begin{align}
\notag
\mathcal{C}^{\text{Precision}}_{\mathcal{M}, \mathcal{E}_{\text{FAct}}}(\ell) \approx \frac{1}{|\mathcal{X}|} \sum_{x \in \mathcal{X}} \mathcal{E}^{\text{Precision}}_{\text{FAct}}\left( \hat{r}_{\ell}(x) , r_{\texttt{en}}(x) \right)
\end{align}
While FActScore is originally designed to measure factual precision, relying solely on precision is insufficient in our cross-lingual consistency setting. A model that abstains from answering in another language, or responds with vague, easily verifiable statements, may achieve high precision despite omitting key information present in the English response (see the example in~\cref{fig:intro_motivating_example}, where the Hindi response omits large amount of information). To properly evaluate consistency, we should also recall--i.e., whether the non-English response captures the full set of relevant information conveyed in English output. As FActScore is asymmetric\,---\,extracting claims from the first argument and verifying them against the second\,---\,we obtain a recall-oriented variant of consistency by simply swapping the response and reference positions:
\begin{equation}
\notag
\mathcal{C}^{\text{Recall}}_{\mathcal{M}, \mathcal{E}_{\text{FAct}}}(\ell) 
= \frac{1}{|\mathcal{X}|} \sum_{x \in \mathcal{X}} 
\mathcal{E}^{\text{Recall}}_{\text{FAct}}\left( r_{\texttt{en}}(x),\ \hat{r}_{\ell}(x) \right)
\end{equation}
In our experiments, we report the harmonic mean  of these two measures to get our final information consistency metric in the form of an F-score.

\subsection{Empathy Consistency}
\paragraph{Motivation.} 
LLMs are increasingly being used in mental health contexts\,---\,from providing therapeutic feedback~\citep{sharma2023human, hsu2025helping} and conducting diagnostic assessments~\citep{raihan-etal-2024-mentalhelp, mohammadi-etal-2024-welldunn} to offering direct support in mental health therapy~\citep{gilbert2024despite, mclennan2025young}. In such high-stakes applications, it is essential that these systems behave consistently and predictably for users across diverse linguistic and cultural backgrounds. A particularly important property in this regard is the ability to express comparable levels of empathy across languages\,---\,both for ensuring equitable user experience and for upholding ethical standards in emotionally sensitive interactions. In this subsection, we introduce the notion of cross-lingual empathy consistency by leveraging the EPITOME framework~\citep{sharma-etal-2020-computational} for measuring expressed empathy in LLM-human conversations. 
%
%
%
\paragraph{The EPITOME framework.}
The EPITOME framework characterizes expressed empathy through three mechanisms: a) Emotional~Reactions~(\texttt{ER}): an expression of emotions~(e.g., warmth, compassion, etc.), b) Interpretations~(\texttt{IP}): an expression of acknowledgment of the user's predicament, and c) Explorations~(\texttt{EX}): an expression of intent to explore user's feelings in further detail.\footnote{We point the reader to \citet{sharma-etal-2020-computational} for a detailed description of the mechanisms and examples.} 
Originally, each LLM response is evaluated for each of these three mechanisms on a three-point Likert scale. We follow~\citet{gabriel-etal-2024-ai} and condense this three-point scale (no, weak, and strong communication) into a binary categorization by merging the `weak' and `strong' communication categories. This means that we are only concerned with the absence or presence of empathetic communication in terms of the three mechanisms.
\paragraph{Notation and setup.} Unlike information consistency where the FActScore-based evaluator directly compares two texts to produce a consistency score, the empathy evaluator first  independently computes the empathy profile of each input, and then compares them to produce an empathy consistency score. 
Consider the three binary empathy classifiers $f_m$ for $m \in \{\texttt{ER}, \texttt{IP}, \texttt{EX}\}$, each trained to detect a specific empathy mechanism in a given response. For any response $r$, we define its \textit{empathy profile} as a binary vector:
\[
\mathbf{e}(r) = [f_{\texttt{ER}}(r),\ f_{\texttt{IP}}(r),\ f_{\texttt{EX}}(r)] \in \{0,1\}^3.
\]
We define the empathy evaluator, $\mathcal{E}_{\text{Emp}}$ as:
\begin{align}
    \small
    \notag
    \mathcal{E}_{\text{Emp}}(r_{\texttt{en}}(x), \hat{r}_{\ell}(x)) = \begin{cases}
    1, & \text{if } \mathbf{e}(r_{\texttt{en}}(x)) = \mathbf{e}(\hat{r}_{\ell}(x)) \\
    0, & \text{otherwise}
    \end{cases}
\end{align} 
We then use~\cref{eq:consistency} to compute the empathy consistency score.

\section{Experimental Setup}
\label{sec:experiment_setup}
In this section, first we will describe experimental details like the LLMs evaluated, languages considered, etc. Following that, we will discuss the specific details for each of the two cross-lingual consistency metrics we use.
\paragraph{LLMs Evaluated.}
We evaluate several open-weight as well as proprietary LLMs at different parameter scales. Specifically, among the open-weight modes, we evaluate Qwen-2.5~\cite{yang2024qwen2} at 14B, 72B parameter scales, Llama-3 at 8B and 70B scales~\cite{grattafiori2024llama}~\footnote{We use Llama-3.1 for the 8B model, and Llama-3.3 for the 70B model. Llama-3.3 does not have a corresponding 8B model and its 70B-variant is known to be much superior to the Llama-3.1 counterpart.}, Gemma-2~\cite{gemma_2024} at 9B and 27B, and Aya-expanse at 8B and 32B~\cite{dang2024aya}.
For proprietary models, we consider GPT-4o~\cite{openai2023gpt4} --both mini and standard size~\footnote{We use \text{gpt-4o-mini-2024-07-18} and \text{gpt-4o-2024-08-06}.}--along with Gemini-2.0-Flash and Flash-Lite~\cite{team2023gemini}.
\paragraph{Languages.} For both of the two consistency evaluations, we evaluate models on 30 languages with different scripts and covering several language families. We perform evaluation on: Arabic (\texttt{ar}), Bengali (\texttt{bn}), Czech (\texttt{cs}), German (\texttt{de}), Greek (\texttt{el}), Spanish (\texttt{es}), Persian (\texttt{fa}), French (\texttt{fr}), Gujarati (\texttt{gu}), Hindi (\texttt{hi}), Indonesian (\texttt{id}), Italian (\texttt{it}), Hebrew (\texttt{iw}), Japanese (\texttt{ja}), Kannada (\texttt{kn}), Korean (\texttt{ko}), Malayalam (\texttt{ml}), Dutch (\texttt{nl}), Punjabi (\texttt{pa}), Polish (\texttt{pl}), Portuguese (\texttt{pt}), Romanian (\texttt{ro}), Russian (\texttt{ru}), Tamil (\texttt{ta}), Telugu (\texttt{te}), Thai (\texttt{th}), Turkish (\texttt{tr}), Urdu (\texttt{ur}), Vietnamese (\texttt{vi}), Chinese (\texttt{zh}).
\begin{table}[]
\centering
\resizebox{\columnwidth}{!}{
\begin{tabular}{@{}lr|lr@{}}
\toprule
Language & Consistency & Language  & Consistency \\
\midrule
Arabic   & 0.90     & Malayalam & 0.90     \\
Czech    & 0.91     & Persian   & 0.89      \\
French   & 0.93     & Russian   & 0.94     \\
Hindi    & 0.91     & Spanish   & 0.94     \\
Japanese & 0.86     & Tamil     & 0.86    \\ \bottomrule
\end{tabular}
}
\caption{Self-consistency and Translationese. We measure information consistency between English and back-translated responses from each language. See~\cref{subsec:setup:ic} for more details.}
\label{tab:automatic_info_consistency}
\end{table}
\subsection{Information Consistency}
\label{subsec:setup:ic}
\paragraph{Prompts.} The authors of this paper manually curate a list of 150 information-seeking prompts from eight different categories from MMLU~\citep{hendrycks2020measuring}, including categories such as college chemistry, African history, etc.
We specifically focus on questions where answers are long-form and should elicit factual information, excluding non-information seeking prompts like ``How are you?'', those based on person opinion like ``What cities have the best nightlife?'', etc. We show an example question and its translations in~\cref{table:disc_ex_full_en,table:disc_ex_full_id,table:disc_ex_full_ko}.
\paragraph{Evaluator.} 
The FActScore evaluator operates in two stages: it first extracts atomic pieces of information from the generated response, and then verifies each claim against the knowledge source (English response in our case). While the original implementation employs GPT-4 for both extraction and verification, relying on a proprietary model is costly at the scale of our experiments. Therefore, we use the open-weight Qwen-2 model at the 14B scale, which we find performs both steps effectively while supporting scalable evaluation.


\paragraph{Translation impact on the evaluator.}
In our information consistency setup, both inputs to the evaluator are in English: the original English response $r_{\texttt{en}}(x)$ and the translated version of a non-English response, $\hat{r}_{\ell}(x)$. To assess the potential impact of translation artifacts\,---\,referred to as translationese in literature~\citep{graham2019translationese}\,---\,on the evaluator, we conduct an automatic validation of the evaluation pipeline. For a subset of languages, we take English responses generated by the \llama 8B model, translate them into each target language using the NLLB model, and then back into English using the same system. We then compute information consistency between the original and back-translated responses, effectively treating this as a self-consistency check. Results in~\cref{tab:automatic_info_consistency} suggest that the Qwen-based evaluator maintains high consistency scores across languages, indicating robustness to translation effects.

\subsection{Empathy Consistency}
\label{subsec:empathy_consistency}
\begin{table}
    \centering
    \begin{tabular}{l | r r}
    \toprule
         \textbf{Mechanism}& \textbf{F1} & \textbf{Accuracy} \\ \midrule
        Emo. Reactions~(\texttt{ER}) & 86.1 &  87.2 \\
        Explorations~(\texttt{EX}) & 92.0 & 95.8  \\
        Interpretations~(\texttt{IP}) & 87.3 &  87.4 \\
        \bottomrule
    \end{tabular}
    \caption{Performance~(in \%) of the evaluator for detecting the three empathy mechanisms.}
    \label{tab:main:evaluator_perf}
\end{table}
\paragraph{Evaluator and prompts.}
We use the publicly-available Reddit subset of the EPITOME dataset~($\sim$3k samples) to train three binary classifiers: $f_{\texttt{ER}}, f_{\texttt{IP}}, f_{\texttt{EX}}$. We sample 100 user utterances from the test split to serve as the evaluation set for computing empathy consistency. These are originally in English, so we translate each of these into the 30 languages. The performance of these classifiers in shown in~\cref{tab:main:evaluator_perf}, their comparison with few-shot GPT-4o, and other details are mentioned in Appendix~\ref{app:emathy_exp_details}.

\begin{table}[t]
\begin{tabular}{@{}lr|lr@{}}
\toprule
Language & Fidelity & Language  & Fidelity \\
\midrule
Arabic   & 4.23     & Malayalam & 4.12     \\
Czech    & 4.13     & Persian   & 4.10      \\
French   & 4.59     & Russian   & 4.28     \\
Hindi    & 4.38     & Spanish   & 4.48     \\
Japanese & 4.14     & Tamil     & 4.16    \\ \bottomrule
\end{tabular}
\caption{Information Fidelity scores for ten languages as judged by native speakers.}
\label{tab:human_eval_info_scores}
\end{table}
\section{Results and Analysis}
\label{sec:results}
Recall one critical assumption in our framework is the fidelity of the translation model we use. As such, we first discuss translation system's fidelity (\cref{subsec:translation_system_fidelty}) before analyzing the cross-lingual consistency evaluation results (\cref{subsec:crosslingual_consistency_results}).


\subsection{Fidelity of the Translation System}
\label{subsec:translation_system_fidelty}
Consider the case of information consistency, where we rely on translating non-English responses into English to enable evaluation with an English-centric metric. As previously noted (\cref{sec:framework}), we assume that the translation process preserves the underlying property of interest--namely, the factual information of the original response. To assess the validity of this assumption, we conduct a human evaluation of translation fidelity.\footnote{Given page limit, we report details about our human evaluation setup in~\cref{appendix:user_study_details}} Specifically, for a generated response in a given language and its corresponding English translation, annotators rate how well the translated text preserves the original information content, using a 5-point Likert scale (with 5 indicating highest fidelity). This evaluation is performed across a subset of ten languages, and the resulting fidelity scores are reported in~\cref{tab:human_eval_info_scores}. We observe an average rating of 4.26, indicating \textit{consistently high scores across all languages}.

While a similar fidelity assessment for empathy consistency is desirable, specifically, to determine whether the translation preserves the empathy profile of the original response, such evaluations require domain expertise in therapeutic communication. Due to the difficulty of sourcing multilingual clinical annotators, we leave it to future work.
\begin{table*}[ht]
    \centering
    \resizebox{\textwidth}{!}{
        \begin{tabular}{@{}lrrrrrrrrrrrr@{}}
        \toprule
        \multicolumn{1}{l}{\multirow{2}{*}{Model}} & \multirow{2}{*}{Avg.} & \multicolumn{4}{c}{Script}                                                  & \multicolumn{5}{c}{Language Family}                & \multicolumn{2}{c}{Train Data} \\ \cmidrule(lr){3-6} \cmidrule(lr){7-11} \cmidrule(lr){12-13}
        \multicolumn{1}{c}{}                       &                          & Latin                    & Indic                    & PArab & Oth1 & Rom & Slav & IA & Drav & Oth2 & In      & Out      \\
        \midrule
        \rowcolor{lightgray!15}
    \multicolumn{13}{l}{\emph{\textbf{Information Consistency}}}\\ 
        \ayasmall & 
        \cellcolor{lightgray!42}0.42 & 
        \cellcolor{lightgray!59}0.59 & 
        \cellcolor{lightgray!11}0.11 & 
        \cellcolor{lightgray!43}0.43 & 
        \cellcolor{lightgray!48}0.48 & 
        \cellcolor{lightgray!61}0.61 & 
        \cellcolor{lightgray!57}0.57 & 
        \cellcolor{lightgray!17}0.17 & 
        \cellcolor{lightgray!6}0.06 & 
        \cellcolor{lightgray!52}0.52 & 
        \cellcolor{lightgray!56}0.56 & 
        \cellcolor{lightgray!9}0.09 \\
        \ayalarge & 
        \cellcolor{lightgray!49}0.49 & 
        \cellcolor{lightgray!65}0.65 & 
        \cellcolor{lightgray!18}0.18 & 
        \cellcolor{lightgray!51}0.51 & 
        \cellcolor{lightgray!55}0.55 & 
        \cellcolor{lightgray!66}0.66 & 
        \cellcolor{lightgray!64}0.64 & 
        \cellcolor{lightgray!26}0.26 & 
        \cellcolor{lightgray!13}0.13 & 
        \cellcolor{lightgray!59}0.59 & 
        \cellcolor{lightgray!63}0.63 & 
        \cellcolor{lightgray!16}0.16 \\
        \gemmasmall & 
        \cellcolor{lightgray!52}0.52 & 
        \cellcolor{lightgray!60}0.60 & 
        \cellcolor{lightgray!41}0.41 & 
        \cellcolor{lightgray!49}0.49 & 
        \cellcolor{lightgray!53}0.53 & 
        \cellcolor{lightgray!61}0.61 & 
        \cellcolor{lightgray!58}0.58 & 
        \cellcolor{lightgray!44}0.44 & 
        \cellcolor{lightgray!36}0.39 & 
        \cellcolor{lightgray!54}0.54 & 
        -- & 
        -- \\
        \gemmalarge & 
        \cellcolor{lightgray!59}0.59 & 
        \cellcolor{lightgray!65}0.65 & 
        \cellcolor{lightgray!53}0.53 & 
        \cellcolor{lightgray!56}0.56 & 
        \cellcolor{lightgray!58}0.58 & 
        \cellcolor{lightgray!66}0.66 & 
        \cellcolor{lightgray!63}0.63 & 
        \cellcolor{lightgray!55}0.55 & 
        \cellcolor{lightgray!50}0.50 & 
        \cellcolor{lightgray!60}0.60 & 
        -- & 
        -- \\
        \llamasmall & 
        \cellcolor{lightgray!32}0.32 & 
        \cellcolor{lightgray!45}0.45 & 
        \cellcolor{lightgray!14}0.14 & 
        \cellcolor{lightgray!25}0.25 & 
        \cellcolor{lightgray!34}0.34 & 
        \cellcolor{lightgray!48}0.48 & 
        \cellcolor{lightgray!42}0.42 & 
        \cellcolor{lightgray!17}0.17 & 
        \cellcolor{lightgray!10}0.10 & 
        \cellcolor{lightgray!35}0.35 & 
        \cellcolor{lightgray!44}0.44 & 
        \cellcolor{lightgray!28}0.28 \\
        \llamalarge & 
        \cellcolor{lightgray!49}0.49 & 
        \cellcolor{lightgray!59}0.59 & 
        \cellcolor{lightgray!33}0.33 & 
        \cellcolor{lightgray!45}0.45 & 
        \cellcolor{lightgray!53}0.53 & 
        \cellcolor{lightgray!61}0.61 & 
        \cellcolor{lightgray!58}0.58 & 
        \cellcolor{lightgray!39}0.39 & 
        \cellcolor{lightgray!29}0.29 & 
        \cellcolor{lightgray!53}0.53 & 
        \cellcolor{lightgray!58}0.58 & 
        \cellcolor{lightgray!47}0.47 \\
        \qwensmall & 
        \cellcolor{lightgray!39}0.39 & 
        \cellcolor{lightgray!54}0.54 & 
        \cellcolor{lightgray!14}0.14 & 
        \cellcolor{lightgray!34}0.34 & 
        \cellcolor{lightgray!44}0.44 & 
        \cellcolor{lightgray!58}0.58 & 
        \cellcolor{lightgray!50}0.50 & 
        \cellcolor{lightgray!19}0.19 & 
        \cellcolor{lightgray!9}0.09 & 
        \cellcolor{lightgray!46}0.46 & 
        \cellcolor{lightgray!53}0.53 & 
        \cellcolor{lightgray!29}0.29 \\
        \qwenlarge & 
        \cellcolor{lightgray!49}0.49 & 
        \cellcolor{lightgray!62}0.62 & 
        \cellcolor{lightgray!27}0.27 & 
        \cellcolor{lightgray!49}0.49 & 
        \cellcolor{lightgray!54}0.54 & 
        \cellcolor{lightgray!64}0.64 & 
        \cellcolor{lightgray!60}0.60 & 
        \cellcolor{lightgray!35}0.35 & 
        \cellcolor{lightgray!19}0.19 &
        \cellcolor{lightgray!56}0.56 & 
        \cellcolor{lightgray!61}0.61 & 
        \cellcolor{lightgray!42}0.42 \\
        \geminismall & 
        \cellcolor{lightgray!65}0.65 & 
        \cellcolor{lightgray!69}0.69 & 
        \cellcolor{lightgray!61}0.61 & 
        \cellcolor{lightgray!64}0.64 & 
        \cellcolor{lightgray!63}0.63 & 
        \cellcolor{lightgray!70}0.70 & 
        \cellcolor{lightgray!68}0.68 & 
        \cellcolor{lightgray!63}0.63 & 
        \cellcolor{lightgray!59}0.59 & 
        \cellcolor{lightgray!65}0.65 & 
        \cellcolor{lightgray!66}0.66 & 
        \cellcolor{lightgray!62}0.62 \\
        \geminilarge & 
        \cellcolor{lightgray!64}0.64 & 
        \cellcolor{lightgray!67}0.67 & 
        \cellcolor{lightgray!60}0.60 & 
        \cellcolor{lightgray!63}0.63 & 
        \cellcolor{lightgray!62}0.62 & 
        \cellcolor{lightgray!68}0.68 & 
        \cellcolor{lightgray!66}0.66 & 
        \cellcolor{lightgray!61}0.61 & 
        \cellcolor{lightgray!59}0.59 & 
        \cellcolor{lightgray!64}0.64 & 
        \cellcolor{lightgray!65}0.65 & 
        \cellcolor{lightgray!61}0.61 \\
        \gptsmall & 
        \cellcolor{lightgray!62}0.62 & 
        \cellcolor{lightgray!69}0.69 & 
        \cellcolor{lightgray!53}0.53 & 
        \cellcolor{lightgray!60}0.60 & 
        \cellcolor{lightgray!62}0.62 & 
        \cellcolor{lightgray!70}0.70 & 
        \cellcolor{lightgray!67}0.67 & 
        \cellcolor{lightgray!56}0.56 & 
        \cellcolor{lightgray!51}0.51 & 
        \cellcolor{lightgray!63}0.63 & 
        -- & 
        -- \\
        \gptlarge & 
        \cellcolor{lightgray!70}0.68 & 
        \cellcolor{lightgray!70}0.73 & 
        \cellcolor{lightgray!60}0.62 & 
        \cellcolor{lightgray!70}0.67 & 
        \cellcolor{lightgray!70}0.66 & 
        \cellcolor{lightgray!70}0.74 & 
        \cellcolor{lightgray!70}0.72 & 
        \cellcolor{lightgray!70}0.65 & 
        \cellcolor{lightgray!60}0.60 & 
        \cellcolor{lightgray!70}0.68 & 
        -- & 
        -- \\
        \rowcolor{lightgray!15}
    \multicolumn{13}{l}{\emph{\textbf{Empathy Consistency}}}\\ 
        \ayasmall & 
        \cellcolor{lightgray!65}0.65 & 
        \cellcolor{lightgray!74}0.74 & 
        \cellcolor{lightgray!49}0.49 & 
        \cellcolor{lightgray!76}0.76 & 
        \cellcolor{lightgray!65}0.65 & 
        \cellcolor{lightgray!73}0.73 & 
        \cellcolor{lightgray!75}0.75 & 
        \cellcolor{lightgray!59}0.59 & 
        \cellcolor{lightgray!43}0.43 & 
        \cellcolor{lightgray!70}0.70 & 
        \cellcolor{lightgray!72}0.72 & 
        \cellcolor{lightgray!50}0.50 \\
        \ayalarge & 
        \cellcolor{lightgray!67}0.67 & 
        \cellcolor{lightgray!76}0.76 & 
        \cellcolor{lightgray!52}0.52 & 
        \cellcolor{lightgray!76}0.76 & 
        \cellcolor{lightgray!64}0.64 & 
        \cellcolor{lightgray!75}0.75 & 
        \cellcolor{lightgray!79}0.79 & 
        \cellcolor{lightgray!55}0.55 & 
        \cellcolor{lightgray!51}0.51 & 
        \cellcolor{lightgray!70}0.70 & 
        \cellcolor{lightgray!73}0.73 & 
        \cellcolor{lightgray!52}0.52 \\
        \gemmasmall & 
        \cellcolor{lightgray!59}0.59 & 
        \cellcolor{lightgray!66}0.66 & 
        \cellcolor{lightgray!55}0.55 & 
        \cellcolor{lightgray!53}0.53 & 
        \cellcolor{lightgray!53}0.53 & 
        \cellcolor{lightgray!65}0.65 & 
        \cellcolor{lightgray!67}0.67 & 
        \cellcolor{lightgray!57}0.57 & 
        \cellcolor{lightgray!53}0.53 & 
        \cellcolor{lightgray!57}0.57 & 
        -- & 
        -- \\
        \gemmalarge & 
        \cellcolor{lightgray!60}0.60 & 
        \cellcolor{lightgray!64}0.64 & 
        \cellcolor{lightgray!58}0.58 & 
        \cellcolor{lightgray!57}0.57 & 
        \cellcolor{lightgray!56}0.56 & 
        \cellcolor{lightgray!66}0.66 & 
        \cellcolor{lightgray!62}0.62 & 
        \cellcolor{lightgray!56}0.56 & 
        \cellcolor{lightgray!58}0.58 & 
        \cellcolor{lightgray!60}0.60 & 
        -- & 
        -- \\
        \llamasmall & 
        \cellcolor{lightgray!45}0.45 & 
        \cellcolor{lightgray!53}0.53 & 
        \cellcolor{lightgray!36}0.36 & 
        \cellcolor{lightgray!38}0.38 & 
        \cellcolor{lightgray!44}0.44 & 
        \cellcolor{lightgray!55}0.55 & 
        \cellcolor{lightgray!51}0.51 & 
        \cellcolor{lightgray!40}0.40 & 
        \cellcolor{lightgray!29}0.29 & 
        \cellcolor{lightgray!46}0.46 & 
        \cellcolor{lightgray!55}0.55 & 
        \cellcolor{lightgray!42}0.42 \\
        \llamalarge & 
        \cellcolor{lightgray!48}0.48 & 
        \cellcolor{lightgray!55}0.55 & 
        \cellcolor{lightgray!41}0.41 & 
        \cellcolor{lightgray!40}0.40 & 
        \cellcolor{lightgray!49}0.49 & 
        \cellcolor{lightgray!57}0.57 & 
        \cellcolor{lightgray!58}0.58 & 
        \cellcolor{lightgray!41}0.41 & 
        \cellcolor{lightgray!39}0.39 & 
        \cellcolor{lightgray!48}0.48 & 
        \cellcolor{lightgray!54}0.54 & 
        \cellcolor{lightgray!47}0.47 \\
        \qwensmall & 
        \cellcolor{lightgray!60}0.60 & 
        \cellcolor{lightgray!67}0.67 & 
        \cellcolor{lightgray!49}0.49 & 
        \cellcolor{lightgray!62}0.62 & 
        \cellcolor{lightgray!59}0.59 & 
        \cellcolor{lightgray!69}0.69 & 
        \cellcolor{lightgray!66}0.66 & 
        \cellcolor{lightgray!50}0.50 & 
        \cellcolor{lightgray!45}0.45 & 
        \cellcolor{lightgray!63}0.63 & 
        \cellcolor{lightgray!66}0.66 & 
        \cellcolor{lightgray!56}0.56 \\
        \qwenlarge & 
        \cellcolor{lightgray!61}0.61 & 
        \cellcolor{lightgray!67}0.67 & 
        \cellcolor{lightgray!54}0.54 & 
        \cellcolor{lightgray!57}0.57 & 
        \cellcolor{lightgray!59}0.59 & 
        \cellcolor{lightgray!72}0.72 & 
        \cellcolor{lightgray!64}0.64 & 
        \cellcolor{lightgray!59}0.59 & 
        \cellcolor{lightgray!47}0.47 & 
        \cellcolor{lightgray!61}0.61 & 
        \cellcolor{lightgray!65}0.65 & 
        \cellcolor{lightgray!58}0.58 \\
            
        \geminismall & 
        \cellcolor{lightgray!65}0.65 & 
        \cellcolor{lightgray!71}0.71 & 
        \cellcolor{lightgray!63}0.63 & 
        \cellcolor{lightgray!64}0.64 & 
        \cellcolor{lightgray!60}0.60 & 
        \cellcolor{lightgray!68}0.68 & 
        \cellcolor{lightgray!72}0.72 & 
        \cellcolor{lightgray!60}0.60 & 
        \cellcolor{lightgray!63}0.63 & 
        \cellcolor{lightgray!66}0.66 & 
        \cellcolor{lightgray!66}0.66 & 
        \cellcolor{lightgray!63}0.63 \\
        \geminilarge & 
        \cellcolor{lightgray!60}0.60 &
        \cellcolor{lightgray!64}0.64 &
        \cellcolor{lightgray!54}0.54 & 
        \cellcolor{lightgray!61}0.61 & 
        \cellcolor{lightgray!59}0.59 & 
        \cellcolor{lightgray!62}0.62 & 
        \cellcolor{lightgray!62}0.62 & 
        \cellcolor{lightgray!55}0.55 & 
        \cellcolor{lightgray!53}0.53 & 
        \cellcolor{lightgray!62}0.62 & 
        \cellcolor{lightgray!61}0.61 & 
        \cellcolor{lightgray!57}0.57 \\
        \gptsmall & 
        \cellcolor{lightgray!76}0.76 & 
        \cellcolor{lightgray!81}0.81 & 
        \cellcolor{lightgray!72}0.72 & 
        \cellcolor{lightgray!73}0.73 & 
        \cellcolor{lightgray!73}0.73 & 
        \cellcolor{lightgray!83}0.83 & 
        \cellcolor{lightgray!80}0.80 & 
        \cellcolor{lightgray!71}0.71 & 
        \cellcolor{lightgray!72}0.72 & 
        \cellcolor{lightgray!76}0.76 & 
        -- & 
        -- \\
        
        \gptlarge & 
        \cellcolor{lightgray!82}0.82 & 
        \cellcolor{lightgray!86}0.86 & 
        \cellcolor{lightgray!77}0.77 & 
        \cellcolor{lightgray!82}0.82 & 
        \cellcolor{lightgray!81}0.81 & 
        \cellcolor{lightgray!88}0.88 & 
        \cellcolor{lightgray!86}0.86 & 
        \cellcolor{lightgray!77}0.77 & 
        \cellcolor{lightgray!79}0.79 & 
        \cellcolor{lightgray!82}0.82 & 
        -- & 
        -- \\         
        \bottomrule
        \end{tabular}
        }
        \caption{Averaged consistency scores for \textit{information}~(top) and \textit{empathy}~(bottom) consistency. We provide details for language and script groupings in~\cref{sec:lang_group_script} and~\cref{sec:lang_group_langfam}. For each model, we also group languages by if they were explicitly included in the training data or not. We refer to each of the model documentations for this.}
        \label{tab:averaged_scores}
        \end{table*}

\subsection{Consistency Results and Analysis} 
\label{subsec:crosslingual_consistency_results}
We present the evaluation results in~\cref{tab:averaged_scores}, and extend the detailed discussion below.

\paragraph{Stronger proprietary models are generally more consistent.} We observe that almost all the open-weight model~(first eight rows) are less consistent than proprietary models. The best-performing proprietary model~(\texttt{gpt-4o-mini}) is significantly more consistent than its best performing open-weight counterpart~(\texttt{gemma-2-27b}) in open-ended generation (0.68 vs 0.59). The consistency gap is even more acute when observed in the empathy scenario with best close-weight model~(\texttt{gpt-4o}) outperforming the best open-weight model~(\texttt{aya-exp-32b}) (0.82 vs 0.67). 
We hypothesize that the consistency of language models has a positive correlation with their performance, as similar observation also applies for models within the same family, e.g., \texttt{gpt-4o} is more consistent than \texttt{gpt-4o-mini}.

\paragraph{LLMs are less consistent for lower resource scripts like Indic.} In~\cref{tab:averaged_scores}, we present results grouped by language scripts. We divide the languages we consider into four script families: \{\texttt{Latin}, \texttt{Indic}, \texttt{Perso-Arabic}, \texttt{Other}\} (details on the grouping of the languages by script in~\cref{sec:lang_group_script}). Notice that models are the most consistent on languages using the Latin script, compared to lower resource scripts like Indic. We hypothesize this result is a consequence of our choice to use English as the reference language. Further, we find that for \texttt{aya-exp-8b}, consistency substantially drops for languages using the Indic scripts compared to the corresponding Latin script languages (0.11 vs 0.59). This finding emphasizes the need for attention towards certain non-Latin scripts.  

\paragraph{LLM responses in low-resource language families are less consistent to English. } 
Consistency results along language family groups follow the related findings observed across script families. We observe that the consistency gap is particularly acute for Dravidian languages, for example, \texttt{aya-expanse-8b} only scores 0.06 compared to Romance languages' 0.61 in the information consistency setting. 
For empathetic consistency, similar degradations are also observed, though not as significant.
Indo-Aryan languages also show significant degradation.
For example, \texttt{Llama-3.1-8B} gets 0.48$\rightarrow$0.17, 0.55$\rightarrow$0.40 from Romance to Indo-Aryan for information and empathy consistency, respectively.
Details about language grouping by language family is in Appendix~\ref{sec:lang_group_langfam}.

\paragraph{LLMs are more consistent with in-training languages.} We find that LLMs are generally more consistent in their responses for languages that are seen in the training data. The difference in consistency are substantial in the \texttt{Aya} family of models while the \texttt{Gemini} models generalize the most to out-of-training languages. 

\paragraph{Scaling mostly helps.} We consider two model sizes per language family in our experiment setup. An increase in model size almost always translates into better task performance~\citep[e.g.,][]{grattafiori2024llama}. 
We observe that in most cases, larger models are more consistent, but it is not always the case.
For example, information and empathy consistency improves for open-weight models like \texttt{gemma-2} and \texttt{Qwen}, but show minor degradation for \texttt{gemini-2.0}.
We intend to more rigorously evaluate the scaling trends of multilingual consistency in the future.

\paragraph{The need for multi-dimensional consistency evaluation.} We recommend the proposed consistency framework to be used in a multi-dimensional manner. In this paper, we instantiated our framework with two such dimensions in information and empathy. We note a highly consistent model along a certain dimension does not guarantee high consistency across all dimensions. For example, \texttt{aya-exp-8b} displays poor information consistency when compared to other open-weight models. However, it is more consistent than certain proprietary models like \texttt{gemini-2.0} in empathy consistency. A comprehensive, multi-dimensional evaluation of consistency should provide further insight into the specific aspects, where models may behave differently across different languages.

\section{Related Works}
\label{sec:relatedworks}
\paragraph{Leveraging output consistency in NLP. }The application of consistency to enhance and evaluate NLP systems has a rich history. For instance, \citet{alberti-etal-2019-synthetic} use round-trip consistency between a question generator and an answer extractor to generate synthetic question-answer pairs. \citet{li-etal-2019-logic} use softened logical constraints as regularizers to improve performance and consistency of NLI models. In a similar vein, \citet{elazar2021measuring} propose a loss-based method to make models robust to relational paraphrases. 

Recently, \citet{twofailure2024chen} evaluate LLMs for self-consistency when posited with hypotheticals and compositional queries in a few-shot setting. In the cross-lingual setting, \citet{qi-etal-2023-cross} evaluate for cross-lingual consistency of LLMs on factual knowledge. \citet{xing2024evaluating} use an embedding similarity-based metric to evaluate cross-lingual semantic consistency for factual question-answers. Our work differs from these works in that we specifically evaluate for open-ended generation tasks, which may or may not have a correct answer. We point the reader to \citet{novikova2025consistency} for a recent survey regarding consistency in language models.

\paragraph{Cross-lingual evaluation of LLMs.} The most popular way of evaluating and comparing cross-lingual performance of LLMs is to evaluate on multilingual benchmarks consisting of parallel data. Such benchmarks focus on commonsense reasoning~\citep{ponti-etal-2020-xcopa,xi2021xstory}, natural language inference~\citep{conneau2018xnli}, world knowledge~\citep[\textit{inter alia}]{dac2023okapi,singh2024global}, which are close-form tasks where cross-lingual performance can be simply evaluated in terms of their respective metrics~(e.g., accuracy). Instead, our framework aims to evaluate consistency of responses between languages without the assumption of a ground truth.

\section{Conclusions and Future Works}
\label{sec:conclusions}
Cross-lingual evaluation of LLMs is generally non-trivial. It is especially challenging for open-ended generation tasks which do not have one correct answer. In this work, we propose a cross-lingual consistency framework that evaluates an LLM's ability to respond similarly for the same inputs across different languages. We instantiate our framework using two consistency evaluators: \textit{information} and \textit{empathy}, and tested 12 popular models in 30 diverse languages. Our findings indicate that models can be highly inconsistent across languages. The inconsistency is higher in open-weight models, especially, for less-represented language families and scripts. We leave investigations on the disparity of low-resource language response and its mitigation strategies to future work. We invite future LLM creators to include our framework in addition to task performance as an essential barometer for multilingual performance.

\section*{Limitations}

The \textit{Translate then Evaluate} approach is central to our framework. Hence, our framework relies heavily on the availability of high-quality translation systems that can translate from a source language $l$ to English, and vice-versa. Of the 7000 languages in the world, most are not supported by commercial translation systems. Google translate supports 249 languages while, NLLB 54B MoE~\citep{costa2022no} supports 204 languages. The development of translation systems for such languages is its own research area~\citep{haddow-etal-2022-survey}. Currently, our framework is not useful for evaluating for such languages without translation support.

We perform a human study for translation fidelity test in the information consistency as shown in table~\ref{tab:human_eval_info_scores}. A similar study would have been ideal to test translation fidelity for the empathy consistency setting. However, such a study requires annotators with domain expertise which is not always straight-forward~\citep{mehta-srikumar-2023-verifying}.

\section*{Acknowledgements}
We thank the members of the UtahNLP group for their valuable insights. We especially thank Ana Marasovi\'c for her help in clarifying the motivation and the experimental setup used in this work. Ashim Gupta is supported by the Bloomberg Data Science Ph.D. Fellowship. This material is based in part upon work supported by the National Science Foundation under Grant \#2217154. The support and resources from the Center for High Performance Computing at the University of Utah are gratefully acknowledged. This research is also supported by National Artificial Intelligence Research Resource
(NAIRR) Pilot (award NAIRR240144) and the Delta advanced computing and data resource which is supported by the
National Science Foundation (award NSF-OAC
2005572). 

\bibliography{custom}

\appendix
\label{sec:appendix}

\section{Experimental Details and Observations}

When running Gemini models, we sometimes receive a \texttt{RECITATION} error from the Google's API. We found there to be no workaround and therefore use a default string of \texttt{``No answer''} in those cases.\footnote{This is an open bug, and there is no existing solution as documented here: \url{https://github.com/google/generative-ai-docs/issues/257}}

\section{Experimental Details for Measuring Information Consistency}

\subsection{Translation Related}

\begin{enumerate}
    \item As mention in the NLLB paper~\citep{costa2022no}, we observe that translation quality with NLLB mdoels significantly degrades with increase in input length. Therefore, we translate sentence-by-sentence instead of all generated text as one input. We find that for splitting sentences properly across languages, we need to use language specific sentence splitters. \textbf{Describe details of sentence tokenizers}.
    
    \item We use the largest MoE model instead of the smaller dense 3.3B model because anecdotally, it produced fewer degenerate translations (although the reported numbers are comparable -- only 1.1\% drop in chrf++).

    \item We set the translation model to generate greedy generations as the original code base uses greedy decoding\footnote{\url{https://github.com/facebookresearch/fairseq/blob/nllb/examples/nllb/modeling/evaluation/generate_multi.py}}. We also experimented with using sampling instead of greedy and found that it had little to no affect on degeneration.

    \item We also handle degeneration wherever we can detect. Given a translation, we see if there are any repeated phrases or repeated patterns. If they do, we generate a translation by sampling for that particular example with temperature as 0.7, and top\_p as 0.95.
\end{enumerate}

\section{Experimental Details for Measuring Discrepancy in Empathy}
\label{app:emathy_exp_details}
This section deals with the experimental details regarding the psychotherapy (empathy) experiments. In this section, we will describe the selection criteria for the empathy evaluator and the evaluation prompts to measure the discrepancy. 

\subsection{Training and Choosing the Empathy Evaluator}
We rely on prior works~\cite{sharma-etal-2020-computational, gabriel-etal-2024-ai} to motivate our experimental decisions for training the empathy evaluator. Our evaluator detects the presence/absence of the use of an empathy communication mechanism by a respondent for a certain seeker post. Following \citet{sharma-etal-2020-computational}'s EPITOME framework, we evaluate for three communication mechanisms:  emotional reaction~(ER), interpretation~(IP), and exploration~(EX).


\paragraph{Data splits.} Since the authors do not publicly share the original splits, we generated our own train-dev-test split using the ratios~(75-5-20) mentioned in \citet{sharma-etal-2020-computational}. These splits are released with the project code base for reproducibility. Note that we create three train-dev-test splits, one each for a kind of empathy communication mechanism. However, the splits are generated at the seeker prompt level, that is, if a certain seeker prompt occurs in the train split for ER empathy communication detection, it will remain in the train split for IP and EX empathy communication detection. We take this decision to avoid test-time leakage. As a result, stratified sampling to maintain the exact distribution between splits becomes a difficult problem. However, our random splits show a similar distribution across the splits~(Table~\ref{tab:split_dist}). The drastic differences in ratios across the empathy communication mechanisms are not relevant since we train different models for each of these mechanisms.

\begin{table}
    \centering
    \begin{tabular}{l | r r r}
         \textbf{Comm. Mechanism}& \textbf{Train} & \textbf{Dev} &  \textbf{Test}\\ \midrule
        ER & 33.7 & 31.2 & 35.5  \\
        IP & 47.2 & 48.1 & 47.5 \\
        EX & 15.8 & 13.0 & 15.4 \\
    \end{tabular}
    \caption{Proportion of datapoints~(in \%) where a certain empathy communication mechanism was used across the splits for the three empathy mechanisms: emotional reaction~(ER), interpretation~(IP), and exploration~(EX)}
    \label{tab:split_dist}
\end{table}

\paragraph{Models Selection and Hyper-parameter Tuning.} We adapt \citet{sharma-etal-2020-computational}'s codebase for our evaluator training with the above-mentioned splits. \cite{sharma-etal-2020-computational} use a bi-encoder model to detect empathetic communication. For the encoder, we consider the following models: \texttt{RoBERTa-base}, \texttt{RoBERTa-large}, \texttt{ModernBERT-base}, and \texttt{ModernBERT-large}. The original work uses \texttt{RoBERTa-base}. We consider the \texttt{ModernBERT} model family for its faster training and larger context window~(8192 vs RoBERTa's 512). We tune the learning rate for the following values: \{1e-5, 2.5e-5, 5e-5, 7.5e-5, 1e-4\} on the dev split. Similarly, we tune the epochs for the following values: \{5, 10, 15, 20\}. We find that \texttt{ModernBERT-large} for the learning rate 2.5e-5 and ten epochs for ER and IP to be the best configuration. For EX, we find that the same model with 5.5e-5 and five epochs to work the best. We train the final classifier with this configuration on the combined train-dev set. The performance of the final evaluator is mentioned in Table~\ref{tab:evaluator_perf}.

\begin{table}
    \centering
    \begin{tabular}{l | r r}
         \textbf{Comm. Mechanism}& \textbf{F1} & \textbf{Accuracy} \\ \midrule
        ER & 86.1 &  87.2 \\
        IP & 87.3 &  87.4 \\
        EX & 92.0 & 95.8  \\
    \end{tabular}
    \caption{Performance~(in \%) of the evaluator for detecting the use of the three communication mechanisms: emotional reaction~(ER), interpretation~(IP), and exploration~(EX)}
    \label{tab:evaluator_perf}
\end{table}

\paragraph{Comparison with GPT-4o. }We compare this bi-encoder based evaluator against a 9-shot inference with GPT-4o. We find that GPT-4o significantly under-performs when compared to the bi-encoder model. The results are mentioned in Table~\ref{tab:gpt4_comp}. The prompt template will be released with the codebase.

\begin{table*}
    \centering
    \begin{tabular}{l | r r | r r}
        &  \multicolumn{2}{c}{\textbf{F1}} & \multicolumn{2}{c}{\textbf{Accuracy}} \\
         \textbf{Comm. Mechanism}& \textbf{Bi-Encoder} & \textbf{GPT-4o} & \textbf{Bi-Encoder} & \textbf{GPT-4o} \\ \midrule
        ER &  &  66.8 & & 66.8\\
        IP &  &  59.8 & & 61.1\\
        EX &  &  70.1 & & 78.0\\
    \end{tabular}
    \caption{Performance~(in \%) comnparison between GPT-4o and the best-performing bi-encoder model.}
     \label{tab:gpt4_comp}
\end{table*}

\subsection{Choice of Evaluation Prompts}
We choose 100 examples from the test split to serve as the evaluation set for computing \discrepancy. We carefully select the prompts so that they are stratified across the two labels. Note that it does not matter if we stratify as we are only interested in the seeker utterances and labeling is performed in conjunction with the human response to the prompt. However, we choose to stratify to reduce any chance of spurious correlations~\citep{gururangan-etal-2018-annotation}. As explained in the previous subsection, since the stratification is performed at the seeker post level, perfect stratification is not possible. 

\section{Discrepancy Examples}
In this section, we have compiled examples where the model generates disparate information for the same prompt in different languages.

\subsection{Open-ended Generation Example}
\label{sec:open_gen_ex}
An example of generations from GPT-4o for the prompt, ``\textit{What were the roles of women in political leadership across pre-colonial African societies?}'', in English, Indonesian, and Korean are shown in Tables~\ref{table:disc_ex_full_en},~\ref{table:disc_ex_full_id}, and \ref{table:disc_ex_full_ko}, respectively. We observe that response to the prompts in Indonesian and Korean to be much more concise than their English counterpart. Our metric aims to capture this discrepancy of information.

\begin{table*}[h]
\centering
\begin{tabular}{p{3cm} p{12cm}}
\toprule
\textbf{Prompt} & What were the roles of women in political leadership across pre-colonial African societies? \\
\textbf{Generation }& In pre-colonial African societies, women played diverse and often significant roles in political leadership, though the extent and nature of their influence varied by region, culture, and historical period. Key roles included:

1. **Queens and Queen Mothers (Matriarchs):**

   * In several African societies, women held formal authority as sovereigns or co-rulers. For example:

     * **Nzinga Mbande** of Ndongo and Matamba (Angola) was a warrior queen who led resistance against Portuguese colonization in the 17th century.
     * The **Queen Mothers of the Asante** (modern-day Ghana), such as Yaa Asantewaa, were powerful political and military leaders.
     * Among the **Lovedu of South Africa**, the Rain Queen (Modjadji) held spiritual and political significance, believed to control rainfall.

2. **Advisors and Regents:**

   * Women often acted as advisors to kings or served as regents during the minority of a male heir. In many cases, Queen Mothers (e.g., in the Oyo and Benin Kingdoms) had substantial influence in court affairs and succession decisions.

3. **Leaders in Matrilineal Societies:**

   * In matrilineal systems, such as among the Akan and Bemba, lineage and inheritance passed through the mother’s line, giving women and maternal uncles significant political influence in succession and property decisions.

4. **Spiritual and Religious Leaders:**

   * Women often held high-ranking spiritual positions that conferred political influence. For example, female priestesses and diviners (e.g., the Igbo *dibia*) mediated between the spiritual and political realms.

5. **Military Leadership:**

   * In some societies, women served as military commanders or warriors. The **Dahomey Amazons** (Fon people of present-day Benin) were an all-female military regiment under the king’s command, demonstrating institutionalized female power in warfare.

6. **Village and Clan Leadership:**

   * At the local level, women frequently served as heads of lineages, clan leaders, or representatives in councils, particularly in societies with decentralized political structures like the Igbo and Yoruba.

In sum, pre-colonial African societies often recognized women's authority in both formal and informal governance structures, contradicting colonial-era and Western portrayals of uniformly patriarchal systems. Would you like examples from a specific region or society?\\
\bottomrule
\end{tabular}
\caption{GPT-4o generation for the given English prompt.}
\label{table:disc_ex_full_en}
\end{table*}

\begin{table*}[h]
\centering
\begin{tabular}{p{3cm} p{12cm}}
\toprule
 \textbf{Prompt} & Apa peran wanita dalam kepemimpinan politik di masyarakat Afrika pra-kolonial?\\
 \textbf{Generation} & Dalam masyarakat Afrika pra-kolonial, wanita memainkan peran penting dalam kepemimpinan politik, meskipun peran tersebut berbeda-beda antar wilayah dan suku. Beberapa peran utama mereka meliputi:

1. **Pemimpin politik dan militer** – Beberapa wanita memegang kekuasaan tertinggi, seperti *Yaa Asantewaa* dari Kekaisaran Asante (Ghana) yang memimpin perang melawan kolonial Inggris, atau *Amina dari Zazzau* (Nigeria) yang dikenal sebagai ratu pejuang dan penakluk.

2. **Ratu ibu (Queen Mothers)** – Di banyak kerajaan seperti Ashanti dan Buganda, ratu ibu memiliki otoritas signifikan, termasuk pengaruh dalam penunjukan raja, nasihat politik, dan pengelolaan wilayah atau istana kerajaan.

3. **Kepala klan atau suku** – Dalam masyarakat matrilineal seperti suku Akan, wanita tua dalam klan (biasanya nenek atau ibu kepala) sering menjadi penentu garis keturunan, warisan, dan keputusan penting komunitas.

4. **Mediator dan penasihat** – Wanita sering memainkan peran sebagai penasihat raja atau dewan, serta mediator dalam konflik antar-kelompok atau keluarga.

Peran ini menunjukkan bahwa kepemimpinan perempuan bukanlah hal yang asing atau sekunder, melainkan integral dalam struktur sosial dan politik pra-kolonial Afrika.

Apakah kamu tertarik pada wilayah atau kerajaan tertentu di Afrika?
\\
\textbf{Generation$^{trans}$ } & In pre-colonial African societies, women played important roles in political leadership, although these roles varied across regions and tribes. Some of their key roles included:

1. **Political and military leaders** – Some women held supreme power, such as *Yaa Asantewaa* of the Asante Empire (Ghana) who led the war against British colonial rule, or *Amina of Zazzau* (Nigeria) who was known as a warrior queen and conqueror.

2. **Queen Mothers** – In many kingdoms such as Ashanti and Buganda, queen mothers wielded significant authority, including influence over king appointments, political advice, and the management of the royal court or territory.

3. **Clan or tribal chiefs** – In matrilineal societies such as the Akan, the elder women in the clan (usually the grandmother or matriarch) often determined lineages, inheritance, and important community decisions.

4. **Mediators and advisors** – Women often played roles as advisors to the king or council, as well as mediators in conflicts between groups or families.

This role shows that female leadership was not an extraneous or secondary thing, but rather integral to the social and political structures of pre-colonial Africa.

Are you interested in a particular region or kingdom in Africa?\\
\bottomrule
\end{tabular}
\caption{GPT-4o generation for the Indonesian translation of the English prompt: ``\textit{What were the roles of women in political leadership across pre-colonial African societies?}''}
\label{table:disc_ex_full_id}
\end{table*}

\begin{table*}[h]
\centering
\begin{tabular}{p{3cm} p{12cm}}
\toprule
\textbf{Prompt} & 식민지 이전 아프리카 사회에서 여성의 정치적 리더십 역할은 무엇이었나요?\\
\textbf{Generation} & 식민지 이전 아프리카 사회에서 여성은 다양한 방식으로 정치적 리더십을 발휘했습니다. 이들은 단순한 조력자에 그치지 않고, 왕비, 여왕, 여사제, 장로, 또는 군사 지도자로 활약하며 중요한 권력을 행사했습니다. 여성 리더들은 종종 영적 권위와 정치 권력을 겸비하여 지역 공동체의 의사결정, 분쟁 조정, 상징적 통치에 깊이 관여했습니다.

예를 들어:

* \*\*다호메이 왕국(현 베냉)\*\*에서는 여성 전사 집단인 ‘아마존’이 존재했고, 이들은 왕을 호위하고 전쟁에 참여하는 핵심 전투 부대였습니다.
* \*\*아샨티 왕국(현 가나)\*\*에서는 여성 지도자인 '퀸 마더(Queen Mother)'가 국왕을 조언하고 후계자 선정에 중요한 역할을 했습니다.
* \*\*이보족 사회(현 나이지리아)\*\*에서는 여성들이 ‘오무아다’ 또는 ‘움우나’와 같은 여성 단체를 조직해 공동체 내 정의 실현과 여성 권익을 위한 자율적 정치 기구를 운영했습니다.

이처럼 여성은 공동체의 생존과 통치를 위한 핵심 축이었으며, 식민주의 이전의 아프리카에서는 정치 권력과 젠더가 더 유연하게 구성되는 경우도 많았습니다.

궁금한 특정 사회나 인물에 대해 더 알아보고 싶으신가요?
\\
\textbf{Generation$^{trans}$} & In pre-colonial African societies, women exercised political leadership in a variety of ways. They were not just helpers; they were also queens, priestesses, elders, and military leaders, exercising significant power. Often, women leaders held both spiritual and political authority, and were deeply involved in community decision-making, conflict resolution, and symbolic governance.

For example:

* In the Kingdom of Dahomey (now Benin) there was a group of female warriors called the Amazons, who were a key fighting force that guarded the king and fought in wars.
* In the Kingdom of Ashanti (now Ghana) a female leader, the Queen Mother, advised the king and played a key role in selecting a successor.
* In the Igbo community (now Nigeria) women organized women’s groups such as the Omuada or Um’una to operate autonomous political institutions for justice and women’s rights within the community.

In this way, women were the linchpin of the survival and governance of the community, and in pre-colonial Africa, political power and gender were often more flexibly structured.

Interested in learning more about a particular society or person?\\
\bottomrule
\end{tabular}
\caption{GPT-4o generation for the Korean translation of the English prompt: ``\textit{What were the roles of women in political leadership across pre-colonial African societies?}''}
\label{table:disc_ex_full_ko}
\end{table*}

\section{Other Experimental Details}
\label{sec: appendix_results}
We define other useful details related to our experiments and results in this section.

\subsection{Language Grouping by Scripts}
\label{sec:lang_group_script}
The language grouping by scripts as used in~\cref{tab:averaged_scores} are provided in~\cref{tab:language_group_script}.

\begin{table*}[]
    \centering
    \begin{tabular}{c c}
        Script Group & Languages \\\midrule
        Latin &  \texttt{cs,de,es,fr,id,it,nl,pl,pt,ro,tr,vi}\\
        Indic &  \texttt{bn,gu,hi,kn,ml,pa,ta,te}\\
        PArb  &  \texttt{ar,fa,ur} \\
        Oth1  &  \texttt{el,iw,ja,ko,ru,th,zh}\\ \bottomrule
    \end{tabular}
    \caption{Language Groups by script. PArb refers to the Perso-Arabic scripts; Oth1 groups all other scripts.}
    \label{tab:language_group_script}
\end{table*}

\subsection{Language Grouping by Scripts}
\label{sec:lang_group_langfam}
The language grouping by scripts as used in~\cref{tab:averaged_scores} are provided in table~\cref{tab:language_group_lf}.

\begin{table*}[]
    \centering
    \begin{tabular}{c c}
        Language Family & Languages \\\midrule
        Rom &  \texttt{de,es,fr,it,nl,pt,ro}\\
        Slav &  \texttt{cs,pl,ru}\\
        IA  &  \texttt{bn,fa,gu,hi,pa,ur} \\
        Drav  &  \texttt{kn,ml,ta,te} \\
        Oth2  &  \texttt{ar,el,id,iw,ja,ko,th,tr,vi,zh}\\ \bottomrule
    \end{tabular}
    \caption{Language Groups by language families. Rom: Romance; Slav:Slavic, IA:Indo-Aryan, Drav:Dravidian, Oth2:Other language families}
    \label{tab:language_group_lf}
\end{table*}

\subsection{Details about Human Study}
\label{appendix:user_study_details}
\begin{figure*}[ht]
    \centering
    \includegraphics[width=\linewidth]{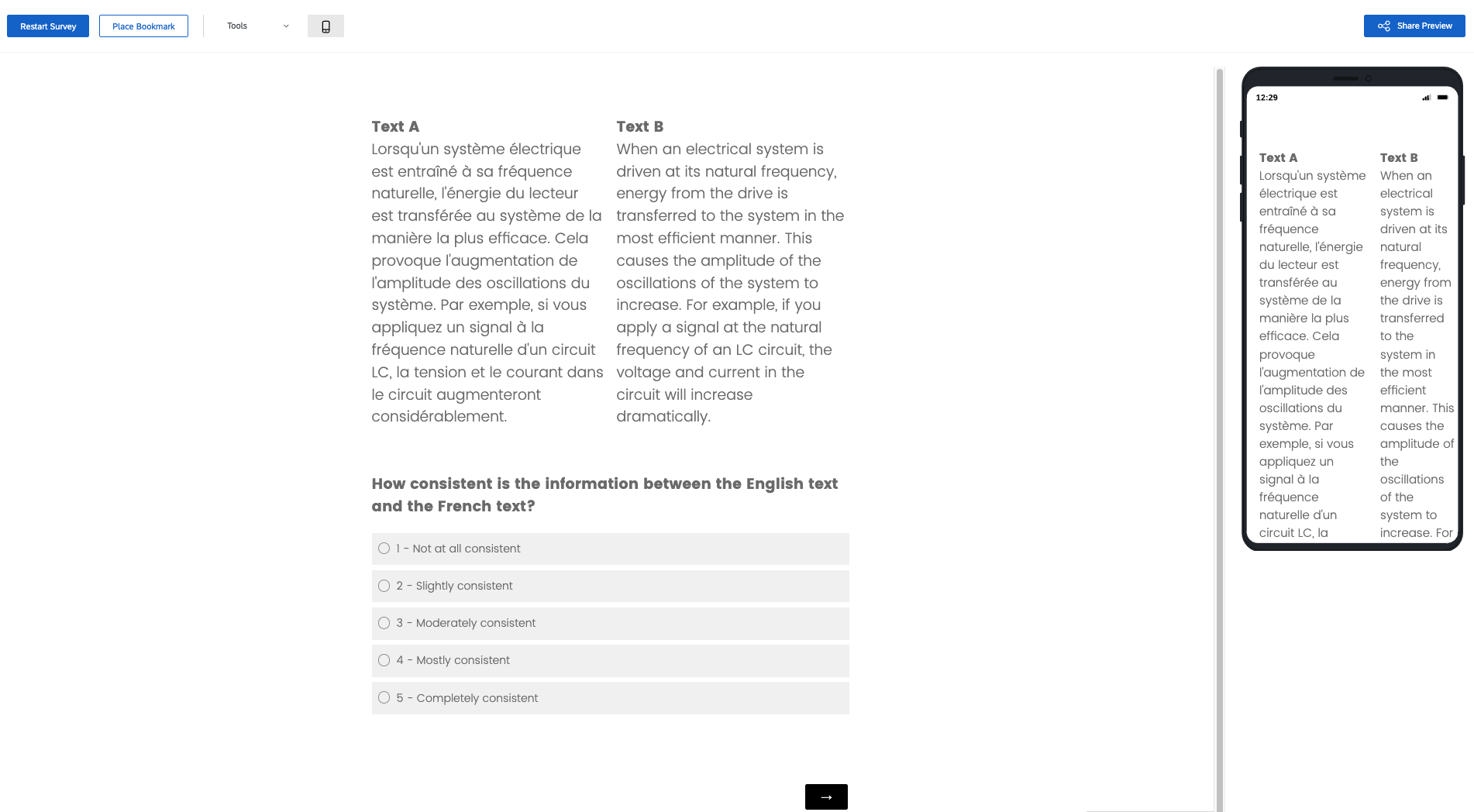}
    \caption{Annotation screen used in the user study. The annotator is asked to rate the translation quality on a Likert scale of 1-5, and higher number indicates better consistency.}
    \label{fig:annotation_screen}
\end{figure*}

For each language, we recruit 10 annotators through prolific and ask each of them to label ten examples each (a total of 100 examples per language). During our pilot study, we found that comparing translations of full long-form responses is a cognitively challenging task. Therefore, we create small snippets of text (usually 1-4 sentences) and present annotators with text in a given language and its corresponding translation. The study was run through qualtrics. 
Each HIT (10 responses) was expected to take a 3-4 minutes, and we paid annotators at the rate of 12\$ per hour (roughly 0.80\$ per HIT). There are some languages for which we finding ten unique annotators would have been hard (the prolific platform provides an estimate of the number of people available for a language). For these languages, we asked annotators to perform multiple HITs with a maximum of 4 per annotator. See~\cref{fig:annotation_screen} for an example annotation screen. 

\subsection{Sentence Tokenizers Used.}
As mentioned before, we translate the multilingual generations one sentence at a time. This necessitates using good, robust sentence tokenizers for each of the 30 languages we considered. For English, German, Spanish, French, Italian, Dutch, Portuguese, Russian, Chinese, Japanese, Korean, Greek, Romanian, Polish, and Indonesian, we used the tokenizers provide by the spacy toolkit. For English and Chinese, we use the \texttt{core\_web\_sm} tokenizers, and for all others we used the \texttt{core\_news\_sm} tokenizers.

For Indian languages like Bengali, Gujarati, Hindi, Kannada, Malayalam, Punjabi, Tamil, and Telugu, we used the tokenizers provided by the IndicNLP toolkit. For Urdu, we used the tokenizer provided by urduhack~\footnote{\url{https://github.com/urduhack/urduhack/blob/master/urduhack/tokenization/eos.py}}.
For Vietnamese, we sed the tokenizer from the \texttt{underthesea} library. 

For all the other languages, we used the Custom tokenizers provided by spacy for each language and use the \texttt{sentencizer} for further rule-based sentence tokenizer. 

We manually verified that each of these sentence tokenizers worked as expected.

\begin{figure*}[ht]
    \centering
    \includegraphics[width=\textwidth]{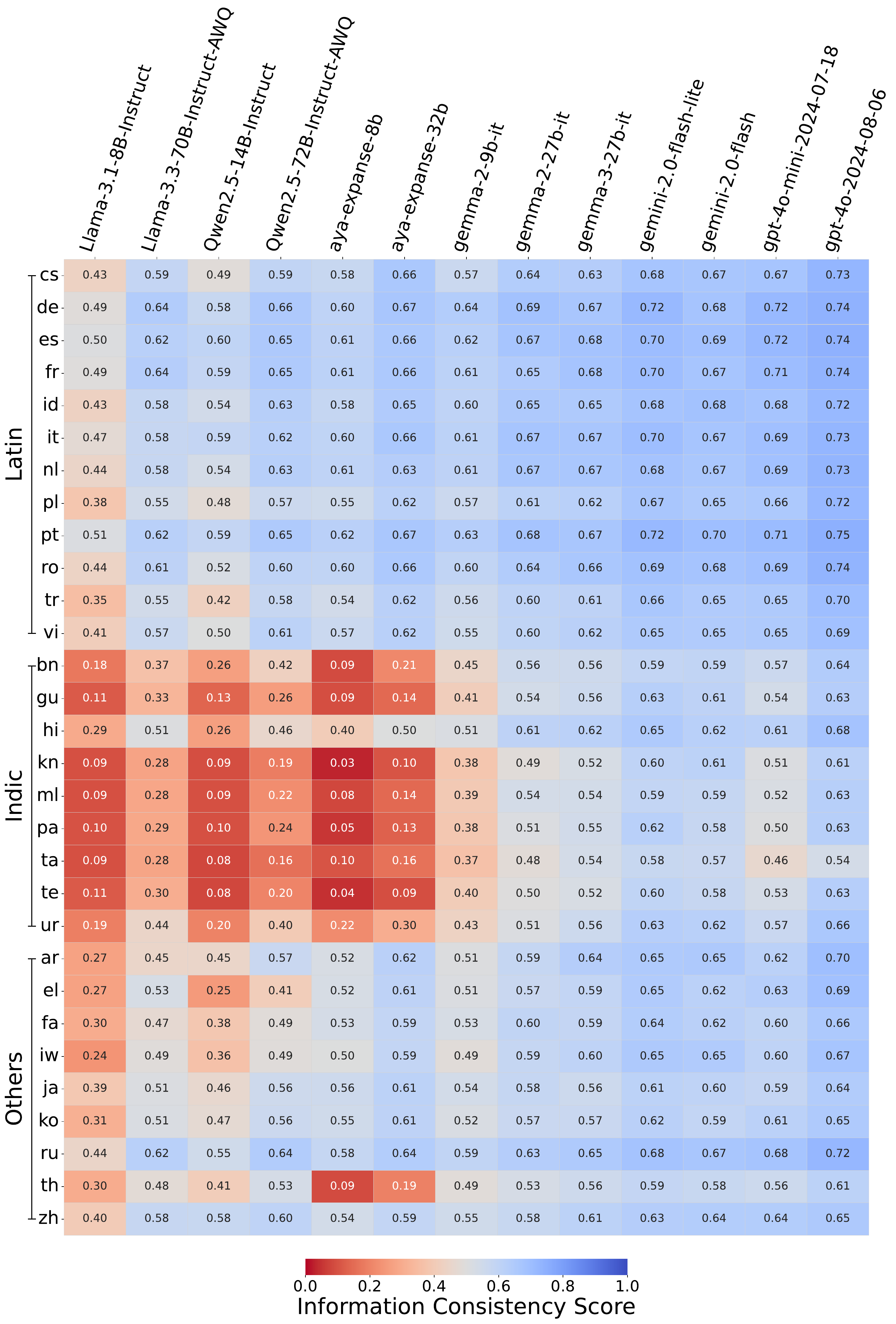}
    \caption{Cross-lingual information consistency scores. The languages are grouped roughly by what they script they below to (Latin, Indic or others).}
    \label{fig:info_consistency_heatmap}
\end{figure*}

\begin{figure*}[ht]
    \centering
    \includegraphics[width=\textwidth]{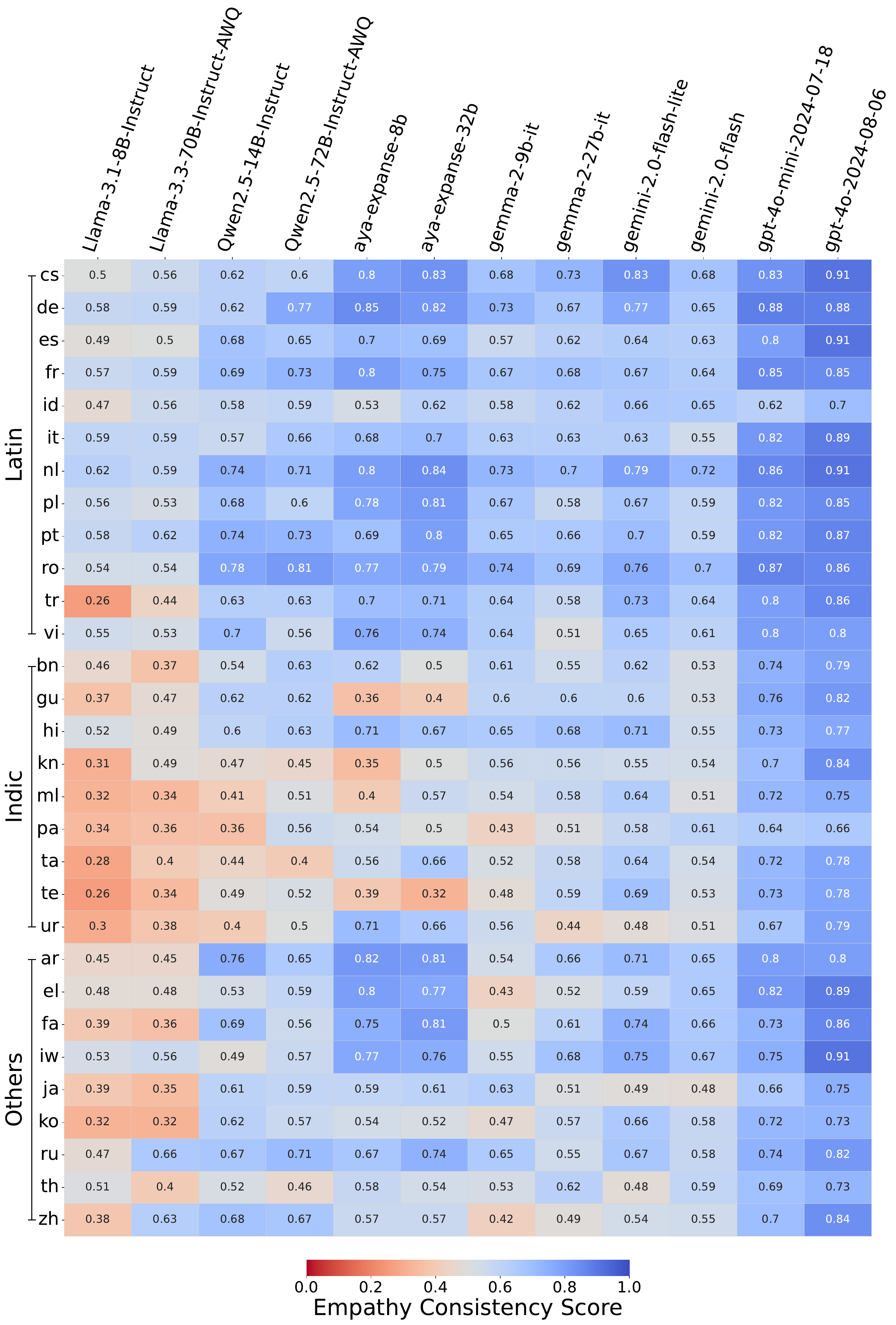}
    \caption{Cross-lingual empathy consistency scores. The languages are grouped roughly by what they script they below to (Latin, Indic or others).}
    \label{fig:info_consistency_heatmap}
\end{figure*}

\end{document}